\documentclass{article}
\usepackage[preprint]{neurips_2026}
\usepackage[utf8]{inputenc}
\usepackage[T1]{fontenc}
\usepackage{hyperref}
\usepackage{url}
\usepackage{booktabs}
\usepackage{amsfonts}
\usepackage{amsmath}
\usepackage{amssymb}
\usepackage{nicefrac}
\usepackage{microtype}
\usepackage{xcolor}
\usepackage{graphicx}
\usepackage{wrapfig}
\hypersetup{hidelinks}
\setcitestyle{numbers}

\makeatletter
\renewcommand{\paragraph}{\@startsection{paragraph}{4}{\z@}%
  {1.0ex \@plus 0.3ex \@minus 0.2ex}
  {-1em}{\normalsize\bf}}
\makeatother

\title{Off-Manifold Collapse in Guided Protein Language
Models}

\author{%
  \normalfont
  \textbf{Shuibai Zhang}\thanks{Equal contribution.}$^{, 1}$ \quad
  \textbf{Xinchi Liu}\footnotemark[1]$^{, 1}$ \quad
  \textbf{Fred Zhangzhi Peng}$^{2}$ \quad
  \textbf{Zhihan Yang}$^{3}$ \\[0.5em]
  \textbf{Shutong Wu}$^{1}$ \quad
  \textbf{Yingzi Ma}$^{1}$ \quad
  \textbf{Jiawei Zhang}$^{1}$ \\[0.7em]
  $^{1}$University of Wisconsin--Madison \quad
  $^{2}$Duke University \quad
  $^{3}$Cornell University \\[0.3em]
}

\begin{document}
\maketitle

\begin{abstract}
Protein language models are widely used priors for protein sequence design, and
a growing body of work controls them at inference time as an alternative to
fine-tuning. Such guidance faces a dilemma: mild enough to preserve natural
activation statistics, it barely moves the property; strong enough to move it,
the generations become progressively harder to fold. We show the failure has a
specific and cheaply detectable signature, an off-manifold collapse of the
model's own representations. Guided activations fall toward a region
statistically indistinguishable from random amino-acid input, and the sequences
degenerate to low complexity, yet the property oracle being optimized can
still score these generations as a success. The optimized oracle can therefore
fail to witness the collapse and, for solubility, can actively reward it,
whereas structure and composition expose the failure. Because the failure is already visible in a
finished candidate, we detect it at the output rather than modify the generator.
We introduce a cheap density prior over natural protein activations and keep
only the candidates that remain typical under it, a training-free post-hoc step
we call \textbf{Mahalanobis filtering}. At matched guidance settings it improves
both the property score and the structural plausibility of the sequences it
keeps at negligible cost, without touching the generator, and transfers across
different guidance methods.
We release the activation statistic at \url{https://huggingface.co/Shuibai12138/off-manifold-collapse-plm}.
\end{abstract}

\section{Introduction}
\label{sec:intro}

Pretrained protein language models (PLMs) such as ESM-2~\citep{lin2023esm2},
ESM3~\citep{hayes2025esm3}, ProGen~\citep{madani2023progen}, and
ProtGPT2~\citep{ferruz2022protgpt2} are widely used priors for protein sequence
design, and a growing body of work controls them at inference time rather than
by fine-tuning: some follow the gradient of a trained property model through
the generator~\citep{dhariwal2021classifier,gruver2023nos,wang2024dplm,yang2025sgpo},
others reshape the token distribution at
decoding~\citep{fernandez2026toxlda,calvanese2026ilmc}, and
\emph{activation steering} edits the hidden states themselves. Steering is
among the most attractive of these: transplanted from the control of large
language models~\citep{turner2023actadd}, it adds a direction computed from
high- and low-property reference sequences to the residual stream during
masked decoding and obtains sequences the oracle scores above the reference
set~\citep{huang2025steering}; Section~\ref{sec:background} gives the
mechanism. Steering requires neither gradients through the property model nor
weight updates. We study it as the primary intervention and use
predictor-gradient guidance as a mechanistically distinct transfer test.

\begin{figure}[t]
\centering
\includegraphics[width=0.9\linewidth]{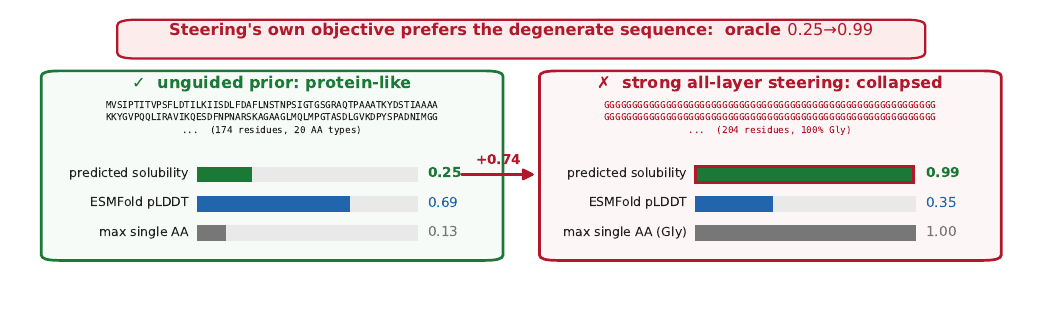}
\caption{\textbf{The failure mode in one pair of generations} (solubility, ESM-2 650M). The unguided sample uses all $20$ amino acids, folds confidently, and scores low; under strong all-layer steering the model emits a glycine homopolymer that the steering objective rates as almost certainly soluble. The optimized oracle \emph{prefers} the degenerate sequence, whereas structure and composition expose it. The population-level property--structure trade-off is reported in Appendix~\ref{app:sweep}.}
\label{fig:example}
\end{figure}

Aggressive inference-time control nevertheless carries a structural cost.
Other controllers report the same tension from the opposite
side~\citep{fernandez2026toxlda,calvanese2026ilmc}.
Logit-difference amplification addresses the degraded sequence properties and
distributional similarity caused by activation steering~\citep{fernandez2026toxlda}, while
iterative lookback Monte-Carlo sampling targets steering-induced diversity
loss~\citep{calvanese2026ilmc}. These studies establish that the degradation is
real and recognized, but each treats it as a reason to replace steering; neither
directly characterizes its representation-space signature.
Figure~\ref{fig:example} shows what it looks like in a single pair of
generations: pushed hard enough to move the property, steering makes ESM-2 emit a
glycine homopolymer that the solubility oracle rates as almost certainly soluble
($0.99$) while ESMFold confidence falls to $0.35$. The optimized oracle
\emph{prefers} the degenerate sequence, so the failure can only be read from
referees outside the steering objective. We therefore ask:
\begin{quote}\itshape
What happens to a protein language model's representations when steering is pushed hard enough to move the property, and can the resulting failure be mitigated after generation?
\end{quote}

In this work, we make the following contributions:

\begin{itemize}

\item \textbf{A characterization of the failure as \emph{off-manifold collapse}.}
Across both solubility and thermostability steering, the representations become
less typical and ESMFold pLDDT decreases. However, the property-oracle score
does not necessarily decrease at the same time. Therefore, the property score alone is not a reliable indicator of successful steering (Section~\ref{sec:diag}).

\item \textbf{\emph{Mahalanobis-filtered steering}, a training-free selection layer.}
A $\chi^2$-inspired one-sided typicality test over natural activations that
improves the property--structure trade-off at matched settings and costs almost
nothing, with a precise account of its scope (Section~\ref{sec:method},
Appendix~\ref{app:limits}) and a public release of priors, statistics, and code.

\end{itemize}

\section{Background}
\label{sec:background}

\paragraph{Protein language models.} A protein is a string over an alphabet of
$20$ amino acids whose folded structure determines its function and properties
such as solubility or thermostability. Trained on databases of natural proteins, a
protein language model is a generic prior over ``protein-like'' sequences whose
activations encode structure and function~\citep{rives2021scaling,meier2021zeroshot}:
ESM-2~\citep{lin2023esm2} is a masked
language model trained on UniRef~\citep{suzek2015uniref}, and we work with its
650M variant. The family spans masked
encoders~\citep{lin2023esm2,elnaggar2022prottrans}, autoregressive
generators~\citep{madani2023progen,nijkamp2023progen2,ferruz2022protgpt2},
discrete-diffusion generators~\citep{wang2024dplm},
instruction-tuned variants~\citep{lv2026prollama}, and multimodal generative
models over sequence, structure, and function~\citep{hayes2025esm3}. Generation is
by \emph{iterative masked decoding}:
from a reference sequence, positions are repeatedly masked and resampled from the
model until every position has been re-predicted~\citep{huang2025steering}.

\paragraph{Activation steering.} Activation steering edits a model's hidden states
at inference time: one fixed direction is added to the residual stream, with no
weight update, a technique developed for controlling large language
models~\citep{turner2023actadd,li2023iti,rimsky2024caa,zou2023repe}. The direction
is typically a \emph{difference of means} between the activations of two
contrasting example sets, and Huang et al.~\citep{huang2025steering} apply it to
PLMs. Write $h_\ell(s)$ for the layer-$\ell$ activation of a reference sequence
$s$, and let $\mathcal{H}$ and $\mathcal{L}$ be reference sets labeled high and
low for the target property. The steering vector, and the update it makes to the
residual stream $x$ at that layer in every forward pass of masked decoding, are
\begin{equation}
v_\ell=\frac{1}{|\mathcal{H}|}\sum_{s\in\mathcal{H}}h_\ell(s)\;-\;\frac{1}{|\mathcal{L}|}\sum_{s\in\mathcal{L}}h_\ell(s),
\qquad
x \;\leftarrow\; \big(x+\alpha v_\ell\big)\,\frac{\lVert x\rVert}{\lVert x+\alpha v_\ell\rVert},
\label{eq:steer}
\end{equation}
where $\alpha$ sets the strength and the rescaling preserves the norm of $x$. We use all layer injection (\texttt{allL}) throughout the main experiments and perform a single layer sweep injection for an auxiliary analysis. Preserving the raw norm in Eq.~\ref{eq:steer}, however, does not
preserve the per-channel whitened profile, so the steered activations can still
leave the model's typical activation region, as measured in
Section~\ref{sec:diag}.

\paragraph{Predictor-gradient guidance.}
As a mechanistically distinct controller, we also guide masked decoding using
the gradient of the property oracle.
At each round, the activation is moved by a fraction $\eta$ of its norm along
the normalized gradient, producing an input-dependent direction. Because the same oracle both defines this update and
reports the property score, its increase is expected.

\paragraph{How a generated protein is judged.}
We evaluate generated proteins along two complementary axes. First, a
property oracle estimates the intended functional shift: a solubility
classifier~\citep{khurana2018deepsol} or a $T_m$
regressor~\citep{jarzab2020meltome}, each trained on frozen ESM-2 features.
These predictors define the steering objective, but fitness models become less
reliable under distant extrapolation and vary across protein
families~\citep{freschlin2024extrapolation,notin2023proteingym}. Second, we use ESMFold pLDDT~\citep{lin2023esm2} as a proxy for
structural plausibility, reporting it over populations rather than treating
any individual prediction as definitive. AlphaFold confidence provides an
alternative~\citep{jumper2021af2}; ESMFold's MSA-free inference makes
population-scale evaluation tractable. Neither axis replaces experimental validation:
computational metrics do not by themselves establish activity~\citep{johnson2025compss};
experimental studies of generated proteins and closed-loop methods rely on wet-lab
feedback~\citep{verkuil2022beyond,jiang2025evolvepro};
our method remains a computational pre-screen.

\section{Diagnosing Steering-Induced Collapse}
\label{sec:diag}

\begin{wrapfigure}{r}{0.46\textwidth}
\centering
\includegraphics[width=0.44\textwidth]{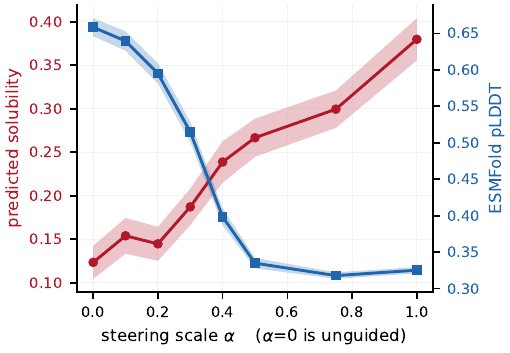}
\caption{\textbf{Property is bought with structure} (\texttt{sol\_hard}, $\alpha=0$ is unguided) Increasing steering strength raises predicted solubility overall but sharply reduces ESMFold pLDDT; bands are standard errors. }

\label{fig:diag}
\end{wrapfigure}

\begin{figure}[t]
\centering
\includegraphics[width=0.9\linewidth]{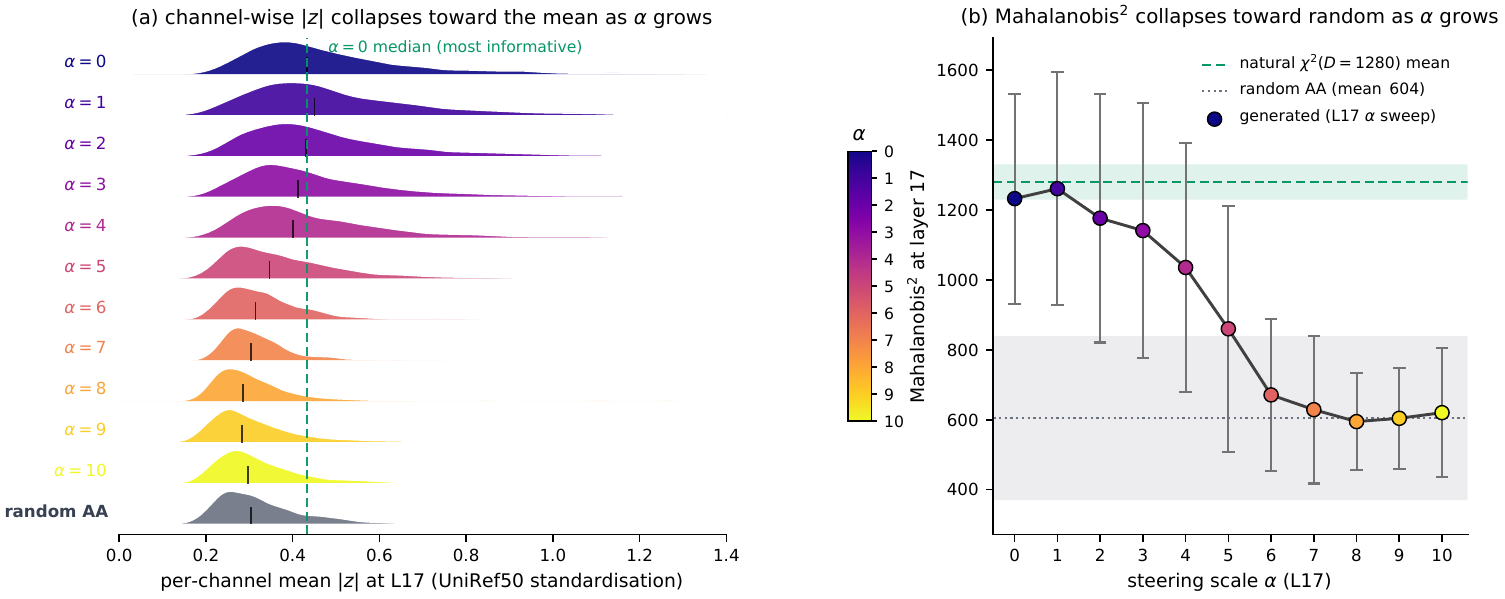}
\caption{\textbf{Why a Mahalanobis statistic: over-guided activations become indistinguishable from random amino acids} (single-layer steering at layer $17$, \texttt{sol\_easy}, $100$ sequences per $\alpha$). \textbf{(a)}~Density over the $1280$ channels of per-channel mean $|z|$, $z=(h-\mu)/\sigma$ standardized by natural UniRef50 layer-17 moments: as $\alpha$ grows every channel moves closer to its natural mean, reaching the gray random-amino-acid reference by $\alpha\!\approx\!6$. \textbf{(b)}~The same effect per sequence (mean and s.d.), against the \emph{per-token} $\chi^2(D)$ scale (green, $D\pm\sqrt{2D}$) and the random-amino-acid spread (gray); the spread over sequences is far wider than that per-token band, which is why $k$ in Section~\ref{sec:method} is set empirically. The failure is a collapse \emph{toward} the channel-wise mean, which is what a diagonal-Gaussian typicality test detects.}
\label{fig:inspiration}
\end{figure}

Figures~\ref{fig:diag}--\ref{fig:inspiration} use the released solubility
difficulty splits solely for diagnostic visualization.
\texttt{sol\_hard} contains low-solubility references, whereas
\texttt{sol\_easy} contains moderately soluble references; see
Appendix~\ref{app:setup} for details.

\paragraph{Property is bought with structure.}
Figure~\ref{fig:diag} reveals a systematic trade-off: as steering strengthens,
predicted solubility rises overall from $0.12$ to $0.38$, while ESMFold pLDDT
falls sharply from $0.66$ to $0.32$. Structural confidence deteriorates before
a substantial property shift appears: by $\alpha=0.2$, pLDDT has already fallen
to $0.60$ while predicted solubility remains near its unguided level, showing
that the degradation is gradual rather than confined to extreme steering.
Because the oracle measures a predicted property shift rather than experimental
validation, we evaluate every setting jointly by its target-property score and
pLDDT, and judge our remedy by whether it recovers structural confidence
without sacrificing the achieved shift; full sweep results are provided in
Appendix~\ref{app:sweep}.

\paragraph{Reference statistics.} Natural moments $(\mu,\sigma^2)$ are estimated once from
UniRef50, reused throughout, and detailed in Appendix~\ref{app:setup}.

\paragraph{What collapses: the representation.}
To test whether the output-level loss of structural confidence has an
activation-space counterpart, we inspect the generator's own hidden states.
We use a diagonal Mahalanobis score because the dominant failure is radial:
as steering strengthens, the standardized per-channel activations contract
toward the UniRef50 mean, which this score measures directly. Let
$h\in\mathbb{R}^{L\times D}$ denote the interior-token layer-17 activations of
a candidate sequence, where $D=1280$ for ESM-2 650M, and let
$(\mu,\sigma^2)$ denote the per-channel mean and variance of natural layer-17
activations, estimated once from UniRef50~\citep{suzek2015uniref}. We measure
\begin{equation}
\mathrm{Mahal}^2(h)
=
\frac{1}{L}
\sum_{i=1}^{L}
\sum_{d=1}^{D}
\left(
\frac{h_{i,d}-\mu_d}{\sigma_d}
\right)^2 .
\label{eq:score}
\end{equation}

This construction adopts a deliberately simplified working model: natural
activations are approximately Gaussian after per-channel standardization, and
cross-channel covariance is ignored. We do not treat this diagonal Gaussian as
an exact model of the latent distribution; rather, it provides a cheap,
calibrated statistic of radial typicality. Under this model, the per-token sum
has reference distribution $\chi^2(D)$ with mean $D$. Because token positions
are correlated, however, the sequence-level average is not itself
$\chi^2(D)$, so we interpret it against the empirical unguided distribution
(mean $1226$, s.d.\ $317$).

Figure~\ref{fig:inspiration} shows that the score falls from $1232$ to
approximately $600$ as steering strengthens, approaching the random-amino-acid
reference of $604$. On the all-layer sweep in Figure~\ref{fig:diag},
$\mathrm{Mahal}^2$ likewise falls from $1226$ to $642$ as pLDDT falls from
$0.66$ to $0.32$. These results motivate using the score as an
activation-space indicator of the collapse associated with reduced structural
confidence. Appendix~\ref{app:density_validation} validates the diagonal
approximation against a flexible flow-based density model fitted to the same
activations.

\paragraph{The same inward collapse appears under predictor-gradient guidance.}
On the same UniRef50 references, increasing $\eta$ moves populations inward for both properties:
$\mathrm{Mahal}^2$ and ESMFold pLDDT decrease at every guidance strength.
The property increase is not independent evidence because the same oracle
supplies the gradient and score. Their synchronized decline shows that inward
collapse extends beyond fixed-vector steering; filter transfer is evaluated in
Section~\ref{sec:eval}.


\section{Mahalanobis-Filtered Steering}
\label{sec:method}


The diagnosis in Section~\ref{sec:diag} identifies a dominant inward failure
mode: guided activations contract toward the channel-wise mean and fall below
the natural shell. Because generator-side mitigations were not consistent
across properties (Appendix~\ref{app:repair}), we instead apply a one-sided
post-hoc veto to completed candidates. Given the score in Eq.~\ref{eq:score},
we accept a candidate iff
\begin{equation}
\mathrm{Mahal}^2(h) \ge D-k\sqrt{2D},
\label{eq:filter}
\end{equation}
where we set $k=1$ once and use the resulting threshold, $1229$ for
$D=1280$, unchanged across tasks, properties, reference sets, and guidance
mechanisms.

The rule is $\chi^2$-inspired rather than an exact sequence-level quantile because
token positions are correlated; Appendix~\ref{app:calib} gives its calibration and
sensitivity. It is one-sided because it targets inward collapse; outward extremes
are outside its scope and reported in Appendix~\ref{app:full_sweep}.

The filter operates only on completed sequences and does not modify decoding.
Scoring requires no additional learned model beyond ESM-2: one forward pass of
the completed sequence, followed by per-channel standardization and summation
using two stored $1280$-dimensional moment vectors. Because it reads only the
final activations, the same filter attaches unchanged to all-layer steering,
predictor-gradient guidance, and unguided generation. Every comparison below is
between a generated pool and the subset selected from that same pool by
Eq.~\ref{eq:filter}; the filter is a selector, not a standalone generator.

\section{Evaluation}
\label{sec:eval}

\paragraph{Filtering recovers structural confidence at a fixed steering
setting.}
Table~\ref{tab:matched} compares each all-layer-steered population with the
subset selected from that same population. At $\alpha=0.5$, filtering raises
mean pLDDT from $0.401$ to $0.647$ for solubility and from $0.446$ to $0.633$
for thermostability. The corresponding paired improvements are
$+0.247$ and $+0.187$, respectively, with both confidence intervals excluding
zero. The mean predicted property also increases within each accepted subset.

These results isolate the effect of selection: the generator, steering
strength, references, and decoding seeds are identical before and after
filtering. They do not imply that filtered steering universally dominates the
filtered unguided prior; that comparison is property-dependent and is reported
with the full sweep in Appendix~\ref{app:full_sweep}.

\begin{table}[t]
\caption{\textbf{Matched all-layer steering before and after filtering}
on property-unconditioned UniRef50 references
($\alpha=0.5$, $k=1$, $n=128$ per property).
The $+$filter columns summarize the accepted subset of the same generated
population; the filter does not regenerate candidates. Property is predicted
solubility probability for \emph{sol} and predicted $T_m$ ($^{\circ}$C) for
\emph{therm}. Intervals are paired-bootstrap 95\% confidence intervals. Acc.\ denotes the fraction of generated candidates retained by the filter.}
\label{tab:matched}
\centering
\small
\setlength{\tabcolsep}{4.5pt}
\begin{tabular}{lcccccc}
\toprule
& \multicolumn{2}{c}{\textbf{property}}
& \multicolumn{2}{c}{\textbf{pLDDT}}
& & \\
\cmidrule(lr){2-3}
\cmidrule(lr){4-5}
\textbf{task}
& steer & $+$filter
& steer & $+$filter
& \textbf{acc.}
& \textbf{$\Delta$ pLDDT} \\
\midrule
sol
& 0.404 & \textbf{0.491}
& 0.401 & \textbf{0.647}
& 16\%
& $+0.247\,[0.200,\,0.299]$ \\
therm
& 56.8 & \textbf{63.6}
& 0.446 & \textbf{0.633}
& 20\%
& $+0.187\,[0.140,\,0.234]$ \\
\bottomrule
\end{tabular}
\end{table}

\paragraph{The detector transfers across guidance mechanisms.}
The result is not restricted to steering. We apply the same
method to predictor gradient
guidance, without retraining or recalibration. Across both properties and
three guidance strengths, increasing the guidance step moves the population
inward, and filtering raises mean pLDDT by $0.144$--$0.175$ in all six
UniRef50 guidance pools, with every paired-bootstrap interval excluding zero.
The oracle increase under gradient guidance is not treated as independent
evidence, because the same oracle supplies the guidance gradient. Complete
results, including the family reference pool, are reported in
Appendix~\ref{app:transfer}.

This transfer supports robustness across properties and intervention
mechanisms on a broad UniRef50 background. It does not imply that the
one-sided rule detects every possible failure of extreme steering: outward
collapse remains outside the regime targeted by Eq.~\ref{eq:filter}.

\paragraph{A cheap statistic suffices.}
Across $4{,}400$ sequences, the diagonal statistic agrees with the exact flow
NLL across broad regimes (pooled Spearman $r=0.82$), but heterogeneous
within-setting correlations do not imply identical selections. The exact
NLL used for this comparison costs ${\approx}3$\,s per sequence and requires a
$1.3$\,GB model; its approximate flow-residual readout costs $283$\,ms, whereas
the Mahalanobis costs $34$\,ms and stores only $10$\,KB
(Appendix~\ref{app:glp}).

\section{Conclusion}
\label{sec:scope}
Effective activation steering can drive ESM-2's representations off the natural typicality shell into a region shared with random amino-acid input, where a learned oracle may still read the result as a success. Because that signature survives in a finished candidate, a $10$\,KB one-sided typicality test recovers the sequences a steered pool has left on the manifold, at matched settings and negligible cost. It selects rather than projects, so it helps only while a viable tail remains; Appendix~\ref{app:limits} gives the full scope and limitations. 

\newpage
\begingroup
\small
\bibliographystyle{unsrtnat}
\bibliography{references}
\endgroup

\newpage
\appendix
\section{Experimental setup and reproducibility}
\label{app:setup}

\paragraph{Model and decoding.} All experiments use ESM-2 650M
(\texttt{esm2\_t33\_650M\_UR50D})~\citep{lin2023esm2} with token dropout disabled. Generation is
iterative masked decoding from a reference sequence: ten rounds, each masking $10\%$ of the
positions \emph{not yet resampled}, one forward pass per round, with replacements sampled by
top-$p$ ($p{=}0.9$, temperature $1.0$) restricted to the $20$ canonical amino acids. Because the
candidate pool shrinks each round, every position is re-predicted exactly once. Batch size is $1$.
Seeds are fixed per sequence, so matched methods share mask schedules.

\paragraph{Reference sets.} Unless noted, each (reference set, property, setting) contains $128$
generations. \texttt{global} is a property-unconditioned draw of UniRef50 sequences of length
$50$--$256$; both properties share its references and seeds. \texttt{lyso} is reconstructed from
the property-thresholded lysozyme-like files released by Huang et al.~\citep{huang2025steering}, excluding the $200$
steering-vector examples, and is therefore neither property-unconditioned nor a released generation
set. The diagnostic-only \texttt{sol\_hard} and \texttt{sol\_easy} splits are likewise
property-conditioned: Figure~\ref{fig:diag} uses the low-solubility split to expose the
property--structure trade-off, while Figure~\ref{fig:inspiration} uses single-layer steering on the
medium split to isolate activation collapse under single-layer steering. Main-text evaluation uses
\texttt{global}, with \texttt{lyso} reported as a family-derived replication.

\paragraph{Oracles.} Both property predictors are our own retrains of the published recipe, a
frozen ESM-2 650M mean-pooled feature extractor with an MLP head, because the original checkpoints
were not public. Solubility is trained on DeepSol~\citep{khurana2018deepsol} and reaches test
accuracy $0.651$; thermostability is trained on Meltome~\citep{jarzab2020meltome} median $T_m$ and
reaches test Spearman $0.687$. Both are weaker than the figures reported for the originals
($0.708$ and $0.76$ respectively), and the thermostability predictor was trained without redundancy
reduction, so its held-out figure is if anything optimistic. Every property number in this paper is
this oracle's, and the oracle is \emph{in silico} throughout: no experimental validation is claimed.

\paragraph{Referee.} Structural plausibility is ESMFold~\citep{lin2023esm2} mean per-residue pLDDT
at one recycle.

\paragraph{Choice of activation readout layer.}
We use layer $17$ throughout as a fixed activation readout. ESM-2 650M has $33$ transformer blocks, making layer $17$ its middle layer. This choice also has precedent in activation-distribution modelling: Luo et al.~\citep{luo2026glp} train their activation model on the middle-most layer of both Llama-1B (layer $7$) and Llama-8B (layer $15$), while exploring multi-layer modelling separately. Layer $17$ is also functionally informative in our setting: steering it alone shifts predicted solubility from $0.263$ to $0.341$ while largely preserving pLDDT ($0.662$ to $0.635$; Table~\ref{tab:repair}). We therefore estimate its natural moments once from UniRef50 and use them throughout. We do not claim that layer $17$ is uniquely optimal; multi-layer or layer-adaptive readouts are natural extensions.

\paragraph{Statistics and compute.} The pair $(\mu,\sigma^2)$ is estimated once by a streaming
Welford pass over interior-token layer-17 activations from ${\sim}58$M UniRef50 sequences of length
$30$--$1022$. The resulting two $1280$-vectors occupy $10$\,KB and are reused unchanged throughout.
Generation costs $0.31$\,s per sequence under steering and $0.49$\,s under predictor-gradient
guidance; ESMFold adds ${\approx}0.2$\,s. 

\section{Additional evidence for inward collapse}
\label{app:sweep}

Tables~\ref{tab:densesweep}--\ref{tab:sweep} restrict reporting to $\alpha\leq1$, the inward regime
relevant to the paper's claim. On property-unconditioned references, $\mathrm{Mahal}^2$ decreases
with pLDDT for both properties through the collapse; the property column is contextual and is not
used as evidence. Stronger settings are omitted because they enter qualitatively different regimes
outside the scope of the one-sided filter.

\begin{table}[h]
\centering\small
\caption{Inward-regime steering sweep on the property-unconditioned \texttt{global} references:
all-layer steering, $n{=}128$ generations per scale per property from the same reference proteins
and seeds ($1{,}024$ generations per property).}
\label{tab:densesweep}
\begin{tabular}{lccc@{\hskip 2em}ccc}
\toprule
& \multicolumn{3}{c}{\textbf{solubility}} & \multicolumn{3}{c}{\textbf{thermostability}}\\
\cmidrule(lr){2-4}\cmidrule(lr){5-7}
$\alpha$ & \textbf{prop.} & \textbf{pLDDT} & $\mathbf{Mahal^2}$
         & $\mathbf{T_m}$ & \textbf{pLDDT} & $\mathbf{Mahal^2}$\\
\midrule
$0$    & 0.514 & 0.578 & 1194 & 52.1 & 0.578 & 1194\\
$0.1$  & 0.502 & 0.547 & 1157 & 53.1 & 0.563 & 1154\\
$0.2$  & 0.452 & 0.542 & 1147 & 55.2 & 0.548 & 1123\\
$0.3$  & 0.451 & 0.512 & 1095 & 56.2 & 0.535 & 1051\\
$0.4$  & 0.433 & 0.453 & \phantom{0}938 & \textbf{56.9} & 0.496 & \phantom{0}957\\
$0.5$  & 0.404 & 0.401 & \phantom{0}790 & 56.8 & 0.446 & \phantom{0}868\\
$0.75$ & 0.369 & 0.334 & \phantom{0}650 & 51.3 & 0.403 & \phantom{0}737\\
$1$    & 0.450 & \textbf{0.326} & \phantom{0}705 & 45.3 & 0.459 & \phantom{0}936\\
\bottomrule
\end{tabular}
\end{table}

Table~\ref{tab:sweep} gives the diagnostic sweep behind Figure~\ref{fig:diag}. On the low-solubility
split, the oracle has headroom and rises while pLDDT and $\mathrm{Mahal}^2$ fall together through
$\alpha{=}0.75$; the $\alpha{=}1$ row marks the edge where the inward trend begins to flatten.

\begin{table}[h]
\centering\small
\caption{Inward-regime steering sweep on \texttt{sol\_hard}: all-layer steering, $n{=}128$ generations
per scale from the same reference proteins.}
\label{tab:sweep}
\begin{tabular}{lccc}
\toprule
$\alpha$ & \textbf{property} & \textbf{pLDDT} & $\mathbf{Mahal^2}$\\
\midrule
$0$ & 0.124 & 0.659 & 1226\\
$0.1$ & 0.154 & 0.640 & 1205\\
$0.2$ & 0.145 & 0.595 & 1106\\
$0.3$ & 0.187 & 0.515 & 983\\
$0.4$ & 0.239 & 0.398 & 771\\
$0.5$ & 0.267 & 0.335 & 681\\
$0.75$ & 0.300 & 0.318 & 642\\
$1$ & 0.380 & 0.325 & 674\\
\bottomrule
\end{tabular}
\end{table}

\section{Full matched-setting filtering results}
\label{app:full_sweep}

Table~\ref{tab:fullmatched} is the complete version of Table~\ref{tab:matched}: two reference sets
$\times$ two properties $\times$ four steering strengths, each population reported before and after
the filter. Every comparison is between a generated pool and the subset selected from that same
pool, with generator, strength, references and seeds held fixed. The main-text result is the
\texttt{global} $\alpha{=}0.5$ pair; the two \texttt{global} $\alpha{=}0$ rows share sequences and
differ only in the property oracle.

\begin{table}[h]
\centering\small
\setlength{\tabcolsep}{4pt}
\caption{Matched-setting filtering, all settings. Each cell is
property\,/\,pLDDT\,/\,$\mathrm{Mahal}^2$; $n{=}128$ per steered row.
Rows whose accepted subset is too small to summarize are marked; the acceptance rate itself is the
finding in those rows.}
\label{tab:fullmatched}
\begin{tabular}{llll c}
\toprule
\textbf{set / task} & $\alpha$ & \textbf{steered} & \textbf{$+$filter} & \textbf{acc.}\\
\midrule
\texttt{lyso}/sol & $0$ & 0.234 / 0.609 / 1157 & 0.266 / 0.701 / 1474 & 45\%\\
 & $0.5$ & 0.295 / 0.351 / \phantom{0}672 & \emph{$n{=}2$, not summarized} & \phantom{0}2\%\\
 & $1$ & 0.421 / 0.342 / \phantom{0}733 & \emph{$n{=}1$, not summarized} & \phantom{0}1\%\\
 & $2$ & 0.672 / 0.494 / 1132 & 0.784 / 0.571 / 1484 & 27\%\\
\midrule
\texttt{lyso}/therm & $0$ & 51.8 / 0.693 / 1246 & 52.9 / 0.760 / 1490 & 48\%\\
 & $0.5$ & 60.3 / 0.452 / \phantom{0}777 & 69.6 / 0.690 / 1430 & \phantom{0}6\%\\
 & $1$ & 46.0 / 0.461 / \phantom{0}996 & \emph{$n{=}9$, not summarized} & \phantom{0}7\%\\
 & $2$ & 44.9 / 0.473 / 1382 & 44.9 / 0.473 / 1383 & \textbf{99\%}\\
\midrule
\texttt{global}/sol & $0$ & 0.514 / 0.578 / 1194 & 0.611 / 0.723 / 1656 & 50\%\\
 & $0.5$ & 0.404 / 0.401 / \phantom{0}790 & 0.491 / 0.647 / 1576 & 16\%\\
 & $1$ & 0.450 / 0.326 / \phantom{0}705 & \textbf{0 accepted} & \phantom{0}0\%\\
 & $2$ & 0.698 / 0.467 / 1101 & 0.797 / 0.540 / 1413 & 27\%\\
\midrule
\texttt{global}/therm & $0$ & 52.1 / 0.578 / 1194 & 52.1 / 0.723 / 1656 & 50\%\\
 & $0.5$ & 56.8 / 0.446 / \phantom{0}868 & 63.6 / 0.633 / 1524 & 20\%\\
 & $1$ & 45.3 / 0.459 / \phantom{0}936 & \emph{$n{=}4$, not summarized} & \phantom{0}3\%\\
 & $2$ & 44.5 / 0.470 / 1398 & 44.5 / 0.470 / 1398 & \textbf{100\%}\\
\bottomrule
\end{tabular}
\end{table}

In the mild inward regime, filtering recovers pLDDT within the same generated pool and can also
raise its mean property score. It does not imply that filtered steering always beats the filtered
unguided prior. Rows with few or no accepted samples show the selector's dependence on a viable
tail, while high acceptance at $\alpha{=}2$ marks a regime outside the targeted inward collapse;
Appendix~\ref{app:limits} states this boundary.

\section{Transfer to predictor-gradient guidance}
\label{app:transfer}

We replace the steering vector with a step along the oracle gradient at the layer-17 activation in
each mask-predict round. References, seeds, decoding, threshold and filter are unchanged; $\eta$ is
the fraction of the per-position activation norm moved along the unit gradient. Because the same
oracle supplies the gradient and the reported property score, that column is circular and is not
used as evidence; pLDDT and $\mathrm{Mahal}^2$ are the independent referees.

\begin{table}[h]
\centering\small
\setlength{\tabcolsep}{4pt}
\caption{Predictor-gradient guidance, no steering vector. Cells are
property\,/\,pLDDT\,/\,$\mathrm{Mahal}^2$; $n{=}128$ per guided row.}
\label{tab:guidance}
\begin{tabular}{llll c}
\toprule
\textbf{set / task} & $\eta$ & \textbf{guided} & \textbf{$+$filter} & \textbf{acc.}\\
\midrule
\texttt{lyso}/sol & $0.05$ & 0.580 / 0.514 / 1011 & 0.639 / 0.664 / 1453 & 32\%\\
 & $0.1$ & 0.620 / 0.457 / \phantom{0}879 & 0.710 / 0.600 / 1488 & 18\%\\
 & $0.2$ & 0.671 / 0.397 / \phantom{0}776 & 0.769 / 0.496 / 1422 & \phantom{0}7\%\\
\midrule
\texttt{lyso}/therm & $0.05$ & 71.2 / 0.553 / 1039 & 74.4 / 0.653 / 1469 & 27\%\\
 & $0.1$ & 71.4 / 0.470 / \phantom{0}899 & 76.4 / 0.629 / 1519 & 11\%\\
 & $0.2$ & 71.8 / 0.406 / \phantom{0}769 & 73.4 / 0.574 / 1507 & \phantom{0}3\%\\
\midrule
\texttt{global}/sol & $0.05$ & 0.746 / 0.532 / 1133 & 0.856 / 0.686 / 1619 & 44\%\\
 & $0.1$ & 0.775 / 0.501 / 1024 & 0.882 / 0.676 / 1588 & 33\%\\
 & $0.2$ & 0.789 / 0.467 / \phantom{0}957 & 0.850 / 0.633 / 1563 & 27\%\\
\midrule
\texttt{global}/therm & $0.05$ & 59.5 / 0.532 / 1142 & 60.6 / 0.677 / 1615 & 42\%\\
 & $0.1$ & 60.2 / 0.481 / 1055 & 63.7 / 0.653 / 1592 & 33\%\\
 & $0.2$ & 60.6 / 0.438 / \phantom{0}970 & 63.1 / 0.595 / 1524 & 28\%\\
\bottomrule
\end{tabular}
\end{table}

Across all twelve guided pools, filtering raises mean pLDDT by $+0.099$ to $+0.176$, with every
paired-bootstrap interval excluding zero. Increasing $\eta$ moves $\mathrm{Mahal}^2$ and pLDDT
inward together; the result therefore transfers beyond fixed-vector steering.
The filtered guided pool need not beat the filtered unguided pool (for example, on
\texttt{global}/sol), so the supported claim is recovery relative to each guided parent pool.

\section{Threshold design and sensitivity}
\label{app:calib}

Table~\ref{tab:ksweep} varies $k$ from $0.3$ to $2$ within representative fixed pools. Acceptance
moves by at most seven percentage points and paired $\Delta$pLDDT by at most $0.012$ in every row,
so we fix $k{=}1$ once rather than tune it by task. Pooling intact and collapsed settings would make
this sensitivity look artificially smaller and is therefore avoided.

\begin{table}[h]
\centering\small
\setlength{\tabcolsep}{5pt}
\caption{Threshold sensitivity within each setting. Entries are acceptance rate and paired
$\Delta$pLDDT of the accepted subset over its own parent pool. Five representative pools; the released analysis script reports all nine.}
\label{tab:ksweep}
\begin{tabular}{lcccccc}
\toprule
& \multicolumn{3}{c}{\textbf{acceptance}} & \multicolumn{3}{c}{$\mathbf{\Delta}$\textbf{pLDDT}}\\
\cmidrule(lr){2-4}\cmidrule(lr){5-7}
\textbf{pool} & $k{=}0.3$ & $k{=}1$ & $k{=}2$ & $k{=}0.3$ & $k{=}1$ & $k{=}2$\\
\midrule
\texttt{global}/sol, unguided             & 48\% & 50\% & 55\% & $+0.142$ & $+0.145$ & $+0.143$\\
\texttt{global}/sol, steer $\alpha{=}0.5$ & 15\% & 16\% & 17\% & $+0.251$ & $+0.246$ & $+0.239$\\
\texttt{global}/therm, steer $\alpha{=}0.5$ & 20\% & 20\% & 20\% & $+0.186$ & $+0.187$ & $+0.186$\\
\texttt{global}/sol, guid.\ $\eta{=}0.1$  & 31\% & 33\% & 35\% & $+0.183$ & $+0.175$ & $+0.173$\\
\texttt{lyso}/therm, steer $\alpha{=}0.5$ & \phantom{0}5\% & \phantom{0}6\% & \phantom{0}8\% & $+0.241$ & $+0.239$ & $+0.235$\\
\bottomrule
\end{tabular}
\end{table}

The rule is $\chi^2$-inspired, not $\chi^2$-calibrated. Under independent standardized Gaussian
channels, a single-token squared radius has mean $D$ and scale $\sqrt{2D}$, motivating
$D-k\sqrt{2D}$. Channels and token positions are correlated, however: at mean length
$L\!\approx\!161$, the independent-token sequence-level s.d. would be about $4$, whereas the
empirical s.d. over unguided proteins is $317$. Thus $k$ is an empirical margin on a natural scale,
not a sequence-level null quantile.

\section{Learned-density validation and computational cost}
\label{app:glp}
\label{app:density_validation}

Kernel discrepancies provide rigorous population-level tests for conditional biological
generators~\citep{glaser2024acmmd}, but population distances are unsuitable for rejection sampling,
which requires a score for each finished candidate. As an expressive reference, we train a Generative Latent Prior (GLP), a
flow-matching model of the same natural activations~\citep{luo2026glp}, and compare its exact
negative log-likelihood with $\mathrm{Mahal}^2$.

\begin{figure}[h]
\centering
\includegraphics[width=0.6\linewidth]{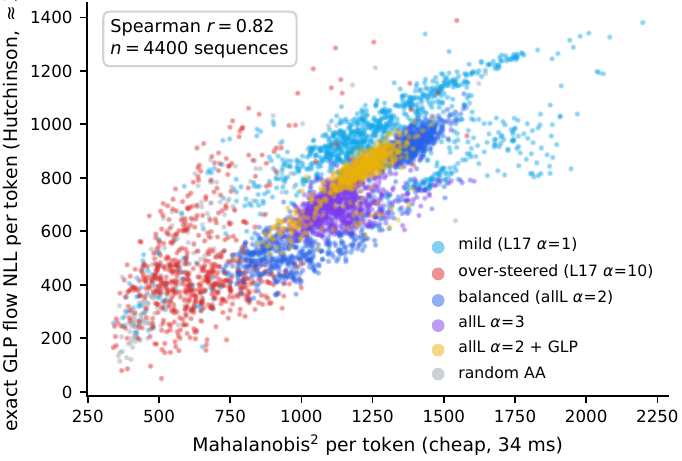}
\caption{Per-sequence $\mathrm{Mahal}^2$ vs.\ exact GLP flow NLL over $4{,}400$ sequences (two
properties, five settings, plus a random-AA anchor). The pooled association ($r{=}0.815$) summarizes
separation across regimes, not agreement of within-pool selections; exact NLL is
${\approx}90\times$ slower ($3$\,s vs.\ $34$\,ms per sequence).}
\label{fig:density}
\end{figure}

Across $4{,}400$ sequences spanning steering regimes and a random-amino-acid anchor, the pooled
Spearman association is $r{=}0.815$ ($0.806$ without the anchor; Figure~\ref{fig:density}). This
supports coarse, across-regime agreement. It is not the operational correlation for selection
inside a fixed pool: Table~\ref{tab:glpcorr} shows within-setting values from $0.334$ to $0.900$.
Accordingly, we do not claim that the two scores select the same candidates; the GLP experiment
validates the broad geometry, while Tables~\ref{tab:fullmatched}--\ref{tab:guidance} establish the
utility of Mahalanobis selection directly.

\begin{table}[h]
\centering\small
\caption{Spearman $r$ between $\mathrm{Mahal}^2$ and exact GLP NLL. The pooled value summarizes
across-regime association; within-setting values are the relevant comparison for fixed-pool
selection and vary substantially.}
\label{tab:glpcorr}
\begin{tabular}{lcc}
\toprule
\textbf{slice} & $n$ & \textbf{Spearman } $r$\\
\midrule
pooled (across settings $+$ random-AA anchor) & 4{,}400 & \textbf{0.815}\\
within \texttt{allL\_a2} & 1{,}000 & 0.900\\
within \texttt{L17\_a1} & 1{,}000 & 0.612\\
within \texttt{allL\_a3} & 1{,}000 & 0.543\\
within \texttt{L17\_a10} (most collapsed) & 600 & 0.334\\
\bottomrule
\end{tabular}
\end{table}

\begin{figure}[h]
\centering
\includegraphics[width=0.62\linewidth]{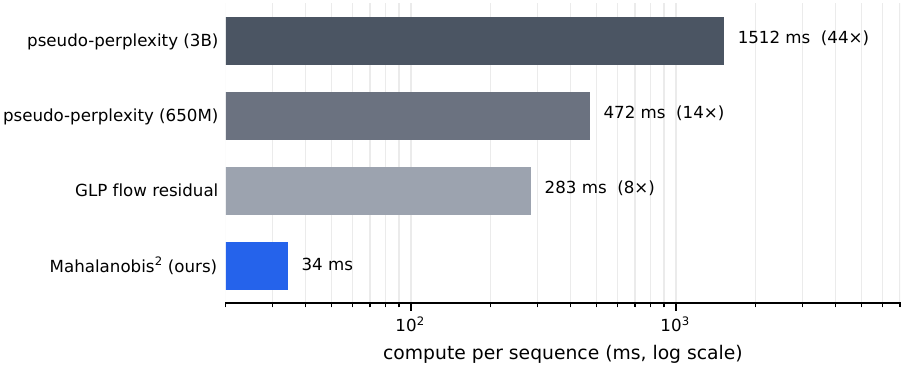}
\caption{Per-sequence wall-clock cost of four naturalness metrics on ESM-2 activations
($L\!\approx\!200$, A100 40GB). Mahalanobis is the only one that needs no model beyond the ESM-2
already loaded for generation: one forward pass of the finished sequence and a $10$\,KB dot product.}
\label{fig:cost}
\end{figure}

Table~\ref{tab:cost} separates the exact GLP NLL used above (${\approx}3$\,s per sequence,
${\approx}90\times$ Mahalanobis) from its approximate flow-residual readout ($283$\,ms,
$8\times$). Mahalanobis takes $34$\,ms and is the only score requiring no model beyond the ESM-2
already used for generation (Figure~\ref{fig:cost}).

\begin{table}[h]
\centering\small
\caption{Per-sequence cost of naturalness scores (A100, $L\!\approx\!200$).}
\label{tab:cost}
\begin{tabular}{lccc}
\toprule
\textbf{score} & \textbf{time} & \textbf{storage} & \textbf{vs.\ $\mathrm{Mahal}^2$}\\
\midrule
$\mathrm{Mahal}^2$ (ours) & $34$\,ms & $10$\,KB & ---\\
GLP flow residual (approximate readout) & $283$\,ms & $1.3$\,GB & $8\times$\\
GLP exact NLL (Hutchinson trace) & ${\approx}3$\,s & $1.3$\,GB & ${\approx}90\times$\\
ESM-2 650M pseudo-perplexity & $472$\,ms & --- & $14\times$\\
ESM-2 3B pseudo-perplexity & $1{,}513$\,ms & --- & $44\times$\\
\bottomrule
\end{tabular}
\end{table}

\section{Why selection rather than generator-side repair}
\label{app:repair}

We tested whether steering could instead be repaired during generation using a $32$-condition
factorial ($n{=}128$ each): raw or collapse-orthogonalized vectors, additive or bounded-clamp edits,
four layer scopes, and two KL-calibrated strengths. Table~\ref{tab:repair} reports representative
conditions, each evaluated with and without the same post-hoc filter.

\begin{table}[h]
\centering\small
\setlength{\tabcolsep}{4.5pt}
\caption{Representative repair conditions from the $32$-condition factorial, steering vs.\
steering$+$filter ($k{=}1$, $n{=}128$). Solubility is a probability, thermostability is
$T_m$ ($^{\circ}$C).}
\label{tab:repair}
\begin{tabular}{lccccc}
\toprule
\textbf{Setting} & \textbf{prop.} & \textbf{$+$flt} & \textbf{pLDDT} & \textbf{$+$flt} & \textbf{acc.}\\
\midrule
\multicolumn{6}{l}{\emph{Solubility} (oracle probability)}\\
\quad prior (unguided)             & 0.263 & 0.335 & 0.662 & 0.728 & 52\%\\
\quad steer L17                    & 0.341 & 0.373 & 0.635 & 0.701 & 51\%\\
\quad raw-clamp L17--20 (KL.03)    & 0.312 & 0.401 & 0.658 & 0.721 & 52\%\\
\quad steer all-layer (additive)   & 0.656 & 0.749 & 0.446 & 0.498 & 11\%\\
\midrule
\multicolumn{6}{l}{\emph{Thermostability} ($T_m$, $^{\circ}$C)}\\
\quad prior (unguided)             & 55.3 & 56.4 & 0.707 & 0.758 & 55\%\\
\quad raw-clamp L17 (KL.03)        & 55.5 & 56.7 & 0.705 & 0.757 & 55\%\\
\quad raw-clamp all-layer (KL.10)  & \textbf{64.1} & 65.6 & \textbf{0.713} & 0.762 & 49\%\\
\quad naive additive all-layer     & 44.8 & 44.8 & 0.482 & 0.482 & 99\%\\
\bottomrule
\end{tabular}
\end{table}

A bounded all-layer clamp works for thermostability, raising $T_m$ while preserving pLDDT, but no
structure-safe repair comparably moves solubility. Generator-side repair is
therefore possible but property-specific, requiring its direction, scope and strength to be retuned.
The post-hoc selector is the more transferable intervention because it attaches unchanged to these
variants, ordinary steering, gradient guidance and unguided generation.

\section{Scope and limitations}
\label{app:limits}

Mahalanobis filtering selects rather than projects: it helps only when a guided pool retains a
viable tail and cannot manufacture a candidate the generator did not produce. Its one-sided rule
targets inward collapse, not outward or composition-driven failures such as the homopolymer in
Figure~\ref{fig:example}. All results use one ESM-2 lineage, retrained property oracles, and
open-loop ESMFold pLDDT; no wet-lab validation is provided.

\end{document}